\documentclass{IOS-Book-Article}

\usepackage[dvipsnames]{xcolor}
\usepackage{mathptmx}
\usepackage{soul}\setuldepth{article}
\usepackage{graphicx}
\usepackage{amssymb}
\usepackage{algorithm}
\usepackage{latexsym}
\usepackage{svg}
\usepackage{algpseudocode}
\usepackage{amsmath}
\usepackage{multirow}
\usepackage{comment}
\usepackage{makecell}
\usepackage[numbers]{natbib}
\usepackage{hyperref}
\usepackage{tabularx}
\usepackage{booktabs}
\usepackage{multirow}
\usepackage{caption}
\usepackage{cleveref}
\DeclareMathOperator*{\argmax}{arg\,max}
\algnewcommand\algorithmicswitch{\textbf{switch}}
\algnewcommand\algorithmiccase{\textbf{case}}
\algnewcommand\algorithmicassert{\texttt{assert}}
\algnewcommand\Assert[1]{\State \algorithmicassert(#1)}%
\algdef{SE}[SWITCH]{Switch}{EndSwitch}[1]{\algorithmicswitch\ #1\ \algorithmicdo}{\algorithmicend\ \algorithmicswitch}%
\algdef{SE}[CASE]{Case}{EndCase}[1]{\algorithmiccase\ #1}{\algorithmicend\ \algorithmiccase}%
\algtext*{EndSwitch}%
\algtext*{EndCase}%
\def\hb{\hbox to 11.5 cm{}}

\begin{document}

\pagestyle{headings}
\def\thepage{}

\begin{frontmatter}              

\title{Few-shot Question Generation for Personalized Feedback in Intelligent Tutoring Systems}

\markboth{}{May 2022\hb}

\author[A,B]{Devang Kulshreshtha}
\author[B]{Muhammad Shayan}
\author[B]{Robert Belfer}
\author[A,D]{Siva Reddy}
\author[B]{Iulian Vlad Serban}
\author[B,C]{Ekaterina Kochmar}
\address[A]{Mila/McGill University, Canada}
\address[B]{Korbit Technologies Inc., Canada}
\address[C]{University of Bath, United Kingdom}
\address[D]{Facebook CIFAR AI Chair}

\begin{abstract}
Existing work on generating hints in Intelligent Tutoring Systems (ITS) focuses mostly on manual and non-personalized feedback. In this work, we explore automatically generated questions as personalized feedback in an ITS. Our personalized feedback can pinpoint correct and incorrect or missing phrases in student answers as well as guide them towards correct answer by asking a question in natural language. Our approach combines cause–effect analysis to break down student answers using text similarity-based NLP Transformer models to identify correct and incorrect or missing parts. We train a few-shot Neural Question Generation and Question Re-ranking models to show questions addressing  components missing in the student's answers which steers students towards the correct answer. Our model vastly outperforms both simple and strong baselines in terms of student learning gains by 45\% and 23\% respectively when tested in a real dialogue-based ITS. Finally, we show that our personalized corrective feedback system has the potential to improve Generative Question Answering systems.
\end{abstract}

\begin{keyword}
Intelligent tutoring systems\sep Natural language processing\sep
Deep learning\sep Question Generation\sep Personalized learning and feedback
\end{keyword}
\end{frontmatter}
\markboth{May 2022\hb}{May 2022\hb}

\section{Introduction}

\begin{table*}
    \centering
    \footnotesize
    \begin{tabular}{|l|l|}
    \hline
    \multicolumn{2}{|l|}{\textit{Exercise Problem:} We want to choose between 2 treatments A and B. For both, we got same mean}\\
    \multicolumn{2}{|l|}{recovery rate but higher variance for treatment A. Which treatment would you discard, and why?}\\
    \hline \hline
        \textit{Student:} Treatment A & \textit{Student:} Treatment A \\
        \hline
    \textit{System [\textcolor{red}{Non-personalized}]:} That's not right. Look & \textit{System [\textcolor{Green}{Personalized}]:} "Treatment A" is correct! \\
    at the variances and provide an explanation why & Try supplying a reason for this idea. Do we\\
    you think one treatment is better than the other. & prefer more homogeneous results or less? \\
    & \textit{Student:} Less \\
    & \textit{System:} Ok, now try to answer original exercise.\\
    \hline
    \textit{Student:} Treatment B? & \textit{Student:} Treatment A, because it is less \\
    & homogeneous than treatment B. \\
    \hline
    \textit{System:} \textcolor{red}{Not really. Let's move to another problem.} & \textit{System:} \textcolor{Green}{That's correct!}\\
    \hline
    \end{tabular}
    \caption{\textcolor{red}{Non-Personalized} vs \textcolor{Green}{Personalized} Feedback Generation in Korbit ITS. The Personalized Feedback pinpoints correct and missing parts in the answer and provides suggestions on how to improve it. In this case, the student forgot to provide reasoning for their answer and is asked a question about the missing part.}
    \label{tab:personalized_vs_non_personalized}
\end{table*}


Intelligent Tutoring Systems (ITS) are AI-powered instructional systems that provide personalized teaching to students \cite{Wenger}. ITS are a low-cost alternative to conventional classroom teaching, and shown to be more effective for tutoring students \cite{st2021comparative, st2022new}. One of the critical aspects of ITS is the ability to provide personalized feedback for exercises. 

Many ITS however rely heavily on expert hand-crafted rules to generate feedback which becomes infeasible for large amounts of educational texts. An important research goal is to thus develop automated feedback systems from student-tutor conversation history \cite{mcbroom2021survey, obermuller2021guiding}. Existing work mainly focuses on non-personalized hints created using template-based methods \cite{blayney2004automated, liu2016automated}. However, students make various type of mistakes (such as grammatical errors, correct answers with incorrect reasoning, and so on), and showing the same hint to address different mistakes is not efficient in improving students' answers, and might even further confuse them. As a result, this can lead to lower motivation and a decrease in the overall study time spent on an ITS platform. 

In this paper, we propose a novel automated personalized feedback system based on deep-learning based Transformer models \cite{zhang2019bertscore, lewis2020bart} to address the above-mentioned problems. Our model first breaks apart student answer into various components by performing cause-effect relation extraction \cite{cao2016role}. Then it matches the components with gold answers using similarity-based Transformers \cite{zhang2019bertscore}, and classifies them into various error categories (such as \textit{missing explanation, incorrect main answer}, and so on). Next, a few-shot Transformer \cite{raffel2020exploring} model generates a personalized natural language question which is combined with the output of the cause–effect analysis to generate question-based feedback. We integrate the feedback in the conversation between an AI-tutor and a student. Such questions are easier to answer compared to the original exercises, as they are aimed at guiding a student towards improving their response.

\Cref{tab:personalized_vs_non_personalized} demonstrates a real interaction with the feedback system. Consider the case where student supplies the correct answer without an explanation. The non-personalized model produces a generic hint irrespective of any student answer, saying \textit{`Thats not right'} even though the main answer is correct. This further confuses the student and causes them to change their correct answer in the next attempt. In contrast, our personalized model first informs the student that their answer is correct and prompts them to supply explanation. It then asks a clarifying question steering the student towards the reasoning part which was missing. As a result, the student is able to provide a correct solution.

We test our method on Korbit ITS,\footnote{https://www.korbit.ai/} a large-scale AI-powered personalized ITS. Students watch video lectures on data science topics and working on problem-solving exercises created by domain experts.
While going through exercises, the student's answers are compared to reference solutions using an ML-based solution checker.
We trigger hint generation when the Korbit's solution checker model marks a student answer as incorrect. We measure the student learning gains after showing our feedback. Our approach outperforms a minimal feedback (simple) baseline by $45\%$ and personalized human feedback (strong) baseline by $23\%$. 

\section{Background: Exercises in Korbit ITS}
Each exercise in Korbit consists of a problem text, and one or more reference solutions. We focus on a particular class of exercises and name them as \textit{cause-effect} exercises. In these exercises, the student is asked about identifying one or several relevant concepts, but they also require to justify the explanation behind their answer. An example of such exercise is \textit{`Can linear regression be applied to classification? Why or why not?'}. Here the expected solution can be decomposed into an \textit{answer (effect)} and \textit{explanation (cause)}. For example, an acceptable solution to the problem above can be \textit{`No, as the output variable of linear regression is continuous'}. Here, the cause is \textit{`The output variable of linear regression is continuous'} and effect is \textit{`No'}. \Cref{tab:cause-effect-examples} illustrates more such examples.

Cause-effect exercises require critical reasoning to solve, as opposed to reading comprehension exercises such as SQuAD \cite{rajpurkar2016squad}. The explanation component can not be usually found directly in any pre-existing knowledge bases or paragraphs. Due to this, the need for personalized feedback is higher in such exercises.

\begin{table}[]
    \centering
    \footnotesize
    \begin{tabular}{c|c}
    \toprule
        \textbf{Connective} & \textbf{Reference solution} \\
        \hline
        because & \textcolor{blue}{It's a discrete variable} \textit{because} \textcolor{Green}{it's counting the number of vehicles} \\
        , & \textcolor{blue}{No}\textit{,} \textcolor{Green}{the feature has 0 weight in the model function.} \\
        then & \textcolor{Green}{If the output is over the threshold} \textit{then} \textcolor{blue}{$x$ is fraudulent} \\
        \bottomrule
    \end{tabular}
    \caption{Decomposition of reference solutions in Korbit ITS into their \textcolor{Green}{cause} and \textcolor{blue}{effect}.}
    \label{tab:cause-effect-examples}
\end{table}

\section{Personalized Feedback Generation Model}
Our model generates feedback in three steps - (i) error classification (ii) Question Generation (iii) Full feedback generation. They are illustrated in \Cref{fig:personalised_feedback_generation} and detailed below:
\begin{figure}
    \centering
    \includegraphics[scale=0.6]{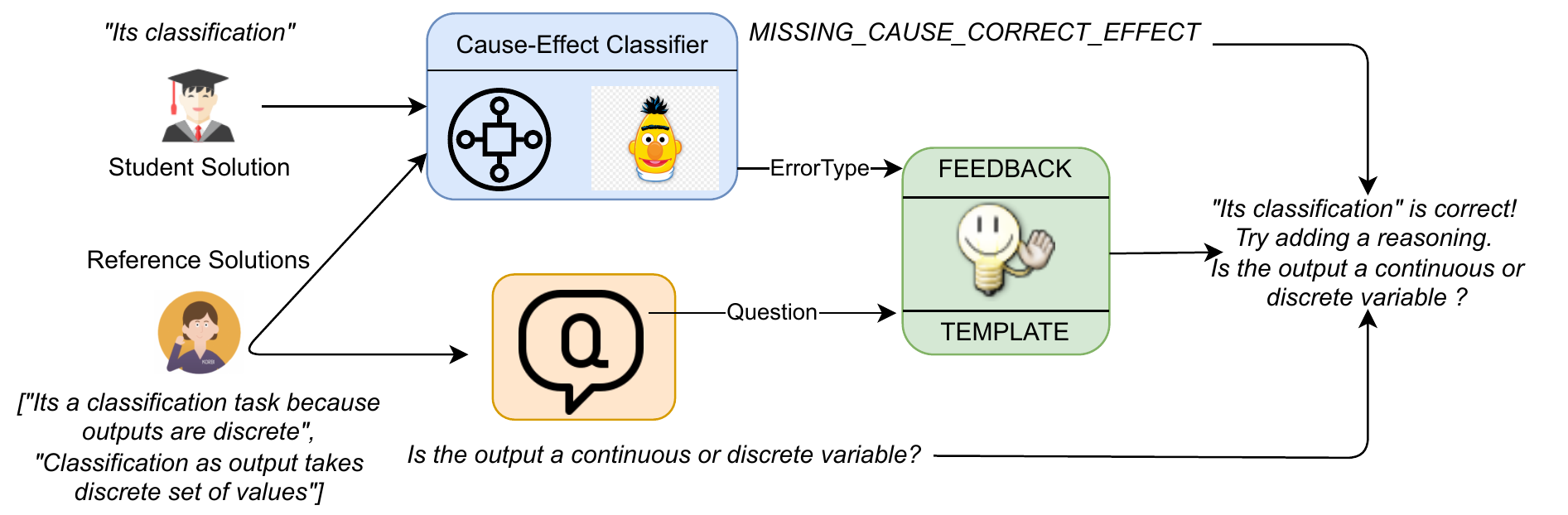}
    \caption{An overview of our personalized feedback generation system: (a) Student solution is classified into its error type using cause-effect extractor and BERT similarity. (b) A few-shot QG model generates question from the \textit{cause} of reference solution. (c) Personalized hint is generated using different feedback templates.}
    \label{fig:personalised_feedback_generation}
\end{figure}
\subsection{Cause-Effect Error Classifier}\label{cause-effect-clf}
\begin{figure}
    \centering
    \includegraphics[scale=0.6]{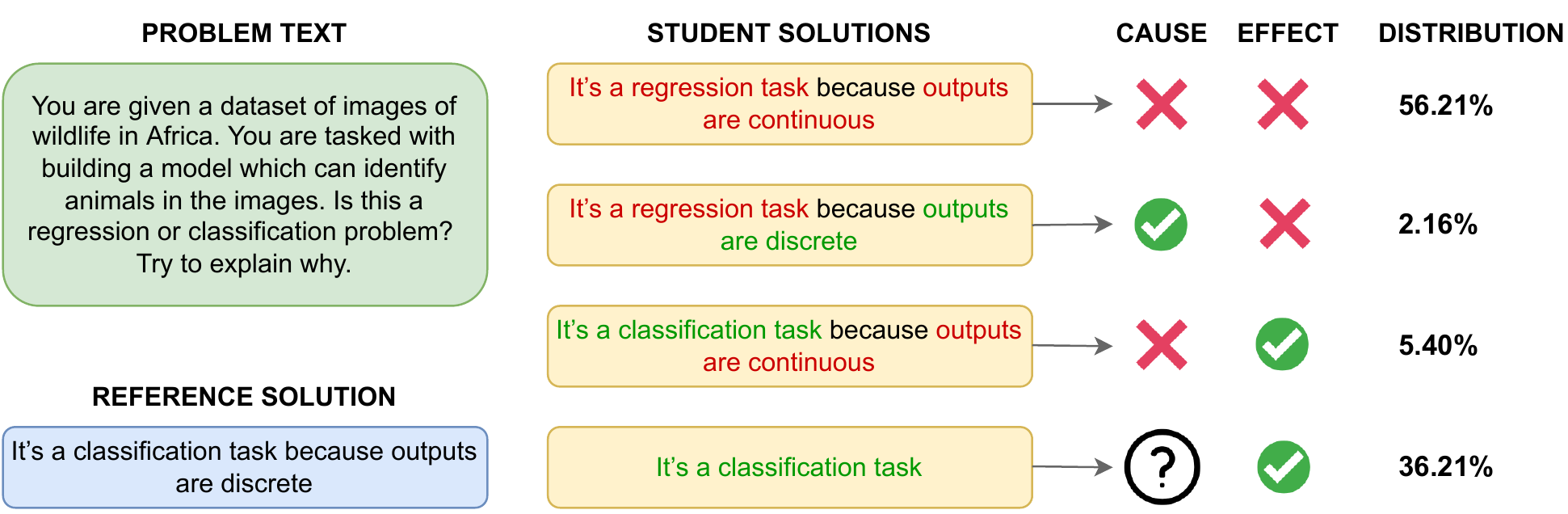}
    \caption{Illustrating various types of student errors for a \textit{cause-effect} exercise in Korbit ITS.}
    \label{fig:error-dist}
\end{figure}

Decomposing a solution into its cause (explanation) and effect (answer) allows classification of student errors. Denote student solution as $s_s \equiv \{c_s, e_s\}$ and gold solution as $s_r \equiv \{c_r, e_r\}$ decomposed into their cause $c$ and effect $e$ by running a cause-effect extractor described in \citet{cao2016role}. The student deficiency falls into one of the four categories:
\begin{itemize}
    \item Incorrect Cause [$c_s \neq c_r$] Incorrect Effect [$e_s \neq e_r$]
    \item Correct Cause [$c_s \equiv c_r$] Incorrect Effect [$e_s \neq e_r$]
    \item Incorrect Cause [$c_s \neq c_r$] Correct Effect [$e_s \equiv e_r$]
    \item Missing Cause [$c_s \equiv \varnothing$] Correct Effect [$e_s \equiv e_r$]
\end{itemize}
\Cref{fig:error-dist} describes examples of all errors for a given exercise, as well as the error distribution generated by running cause-effect extractor over $7,000$ incorrect solutions.

To detect the error type, we match student cause-effect text with reference solution using BERTScore \cite{zhang2019bertscore}. BERTScore uses pre-trained BERT \cite{devlin2019bert} contextualised embeddings and computes overall similarity using weighted mean of cosine similarity between their tokens. It correlates better with human judgments compared with n-gram overlap based metrics (e.g. BLEU, ROUGE etc). BERTScore has been used as an evaluation metric for image captioning \cite{zhang2019bertscore}, summarization (\cite{li2019deep}), machine translation (\cite{unanue2021berttune}) etc. BERTScore returns a score ($0-1$) between student and reference cause/effect. If similarity exceeds a threshold ($=0.8$, set manually) then cause/effect is considered correct.

\subsection{Few-shot Question Generation}\label{question-generation}
Our goal is to generate a question which forces the student to think about the incorrect/missing components in their solutions, and improve their answers. Our QG model pipeline comprises of four steps described below:

\subsubsection{Dataset creation}\label{qg_dataset_creation}
We create a dataset by randomly sampling around 112 cause-effect exercises, giving us around 300 reference solutions for those exercises. We then ask four domain experts to write a question from the reference solutions, giving 75 examples to each annotator. Questions are written to \textit{not} reveal the \textit{effect/answer} and hence created only from \textit{explanation} of reference solution. The annotators mainly write three type of questions - open-ended, binary, and binary with alternatives. Examples of such types are shown in \Cref{tab:question_type_score} (the `Score' column will be explained later in \Cref{question_reranking}). All annotators also annotate a common set of 20 questions to ensure that annotators have low variance in questions. We find the common questions are quite similar and follow the above guidelines.
\begin{table}[]
\footnotesize
    \centering
    \begin{tabular}{c|c|c}
        \hline
        \multicolumn{3}{c}{\textit{\textbf{Reference Solution: } It is classification because coin flip outcome is discrete.}}\\
        \hline
        \textbf{Question Type} & \textbf{Example} & \textbf{Score} \\
        \hline
        Binary & Is flipping a coin discrete? & 0.5 \\
        Binary with alternatives & Is flipping a coin discrete or continuous? & 0.8 \\
        Open-Ended & What kind of action is flipping a coin? & 1 \\
        \hline
    \end{tabular}
    \caption{Taxonomy of questions written by annotators and corresponding scores used for question re-ranking. }
    \label{tab:question_type_score}
\end{table}
\subsubsection{Few-Shot Question Generation (QG) model}
After data collection, we train a QG model to generate questions from reference solutions. We frame QG as a neural sequence-to-sequence task similar to \cite{du2017learning} where an encoder reads input text and decoder produces question by predicting one word at a time. We experiment with two pre-trained Transformers: BART \cite{lewis2020bart} and T5 \cite{raffel2020exploring}.
\paragraph{\textbf{T5}}is an encoder-decoder model pre-trained on a mixture of supervised and unsupervised NLP tasks where each task is converted into text-to-text input-output. T5 works well on a variety of conditional sequence generation tasks such as summarization \cite{rothe2021thorough}, machine translation and question generation \cite{dong2019unified}. We name the model as \textit{T5-QG}.
\paragraph{\textbf{BART}}is a Transformer autoencoder pre-trained to reconstruct text from noisy text inputs. For QG, it learns a conditional probablity distribution $P(q|r)$ to generate question $q$ from reference solution $r$. We experiment with two pre-trained checkpoints - a) original BART-base checkpoint provided by authors and b) BART model trained on 50K MLQuestions dataset using back-training algorithm \cite{kulshreshtha2021back}. The latter model is able to generate good-quality questions for data science domain which is also our domain of interest in Korbit ITS. We denote them as \textit{BART-QG} and \textit{BART-ML-QG}. 

We split the data into 220 train, 40 validation and 40 test examples to train these models. Refer to \Cref{qg_model_training_details} for model training details.
\subsubsection{Improving Question Generation using Re-Ranking}\label{question_reranking}
To improve question quality, we train a question re-ranker to chose the best question: First we generate $k = 3$ questions per reference solution using beam search decoding algorithm \cite{freitag2017beam} for $80$ randomly sampled reference solutions. Then we ask four domain experts to rate the usefulness of $240$ generated questions on a scale of 1-5. Rating is done keeping in mind the \textit{factual correctness, fluency} and \textit{relevance} of question to the input reference solution. Additionally, good quality questions based on question type are given higher score based on the preference - \textit{\{open-ended $>$ binary with alternatives $>$ binary\}} question (see \Cref{tab:question_type_score} for question type examples). The 240 examples are distributed equally amongst three annotators. We find that the mean ratings given by each annotator was quite similar - $3.35, 3.4, 3.46$. Additionally a common set of 20 examples are annotated by all annotators and we record an inter-annotator agreement of $0.75$ which shows substantial agreement according to \citet{landis1977measurement}.

Finally we train a Linear Regression model to predict usefulness taking the reference solution and generated question as input on 200 examples, and test on 40 examples. The input features to the regression model are -
\begin{itemize}
    \item \textbf{Sentence Embeddings:} We use Sentence-BERT \cite{reimers2019sentence} to extract 768 dimensional embeddings from question. The Sentence-BERT uses siamese and triplet network structures to derive sentence embeddings from BERT and have been shown to perform extremely well in common STS tasks and transfer learning tasks \cite{reimers2019sentence}.
    \item \textbf{Well-formedness:} We train a BERT binary classifier to predict whether a question is well-formed or ill-formed on Google Well-formedness dataset \cite{faruqui2018identifying}. We use the well-formedness probability of generated hint question as the \textit{well-formed} feature.
    \item \textbf{Fluency:} We finetune a GPT-2 LM \cite{brown2020language} on the 300 original hand-written questions (\Cref{qg_dataset_creation}) using causal language modeling (LM) objective. The negative of LM perplexity of generated question is used as \textit{fluency} feature.
    \item \textbf{Model Confidence:} This feature is computed as the negative loss of model when the generated question is considered as ground truth.
    \item \textbf{Question Type:} We want to penalise simple questions and reward questions which are more diverse and challenging to answer. We come up with a simple heuristic depicted in \Cref{tab:question_type_score} to compute question type score feature of a question.
\end{itemize}
We get $772$ dimensional feature vector and train our regression model using Ordinary Least Squares (OLS) objective on 200 examples. During inference, we use this question re-ranker to select the best question from the 5-best list for each reference solution. 

After training the Question Generation and reranker model, we generate questions from all 1470 reference solutions in Korbit ITS using above models.

\subsection{Providing Feedback}
Using the output of cause-effect classifier and question generator, we provide feedback to reveal student deficiencies and suggest improvements. First we find the reference solution $s_r$ closest to student solution $s_s$ using BERTScore similarity. Then according to each error category identified by cause-effect classifier in \Cref{cause-effect-clf}, we create feedback described below (full algorithm is described in \Cref{hint_gen_algo})-

\begin{algorithm}[t]
\begin{algorithmic}[1]
\footnotesize
\Require {
Exercise problem $Q$, reference answers $ \mathcal{R} \equiv \{s_r^i\}_{i=1}^m$, incorrect student answer $s_s$, Cause-Effect Extractor $\boldsymbol{\theta_{CE}}$, BERTScore model $\boldsymbol{\theta_{BS}}$, BERTScore similarity threshold $\boldsymbol{\tau_{BS}}$, Question Generator $\boldsymbol{\theta_{QG}}$.
}
\Ensure Personalized hint $h$
\State \textcolor{blue}{/*Find reference answer closest to student answer*/}
\State $sim \leftarrow []$
\For{$s_r \in \mathcal{R}$}
\State add $\boldsymbol{\theta_{BS}}(s_r, s_s)$ to $sim$
\EndFor
\State $s_r \leftarrow \argmax_{i} (sim)$
\State \textcolor{blue}{/*Classify student error and generate personalized hint*/}
\State $(c_r,e_r) \leftarrow \boldsymbol{\theta_{CE}}(s_r); (c_s,e_s) \leftarrow \boldsymbol{\theta_{CE}}(s_s)$ \Comment{Run cause-effect extractor}
\State $q = \boldsymbol{\theta_{QG}}(s_r)$ \Comment{Generate question from reference solution.}
\Switch{[$c_r,e_r,c_s,e_s$]}
    \Case{$c_s \neq c_r \textbf{\ and\ } e_s \neq e_r$ \Comment{[$\theta_{BS}(c_s, c_r) < \tau_{BS} \textbf{\ and\ } \theta_{BS}(e_s, e_r) < \tau_{BS}$]}}
      \State \Return \textit{``$\{e_s\}$ is incorrect. \{$q$\}?''}
    \EndCase
    \Case{$c_s \neq c_r \textbf{\ and\ } e_s \equiv e_r$ \Comment{[$\theta_{BS}(c_s, c_r) < \tau_{BS} \textbf{\ and\ } \theta_{BS}(e_s, e_r) \ge \tau_{BS}$]}}
        \If{$c_s \equiv \varnothing$}
        \State \Return \textit{``\{$e_s$\} is correct! Try supplying a reason for it. \{$q$\}?''}
        \Else
        \State \Return \textit{``\{$e_s$\} is correct! Try changing your reasoning. \{$q$\}?''}
        \EndIf
    \EndCase
    \Case{$c_s \equiv c_r \textbf{\ and\ } e_s \neq e_r$ \Comment{[$\theta_{BS}(c_s, c_r) \ge \tau_{BS} \textbf{\ and\ } \theta_{BS}(e_s, e_r) < \tau_{BS}$]}}
        \State \Return \textit{``Did you mean \{$e_r$\} because \{$c_s$\}?''}
    \EndCase
\EndSwitch
\end{algorithmic}
\caption{Personalized Feedback Generation in Korbit ITS}
\label{hint_gen_algo}
\end{algorithm}

\paragraph{Incorrect Cause [$c_s \neq c_r$] Incorrect Effect [$e_s \neq e_r$]}
First the system reveals error type by saying - \textit{``$\{e_s\}$ is incorrect.''}. Then it asks a question generated from $s_r$ using QG model. Then the student responds to that question, after which we completely ignore their answer. The system then asks the student to answer the original exercise again.

\paragraph{Incorrect Cause [$c_s \neq c_r$] Correct Effect [$e_s \equiv e_r$]}
Since the main answer (effect) is correct, first the system outputs - \textit{``$\{e_s\}$ is correct! Try changing your reasoning.''}. Then similar to previous error type, we ask a sub-question generated from $s_r$. After student answers this sub-question, the interface will ask them to answer original exercise again.

\paragraph{Missing Cause [$c_s \equiv \varnothing$] Correct Effect [$e_s \equiv e_r$]}
We show similar hint as previous error category, saying - \textit{``$\{e_s\}$ is correct! Try supplying  a reason for it. \{$q$\}?''}, where $q$ is the generated question. This example is also illustrated in \Cref{fig:personalised_feedback_generation}.

\paragraph{Correct Cause [$c_s \equiv c_r$] Incorrect Effect [$e_s \neq e_r$]}
In practice this scenario rarely occurs. Since student supplied correct explanation, it is likely that student supplied incorrect answer by mistake. In this case we repair student's solution by asking an MCQ question - \textit{"Did you mean \{$e_r$\} because \{$c_s$\}?"}. The interface supplies two options to chose from - "Yes, I agree" and "No, I disagree". If the student chooses former option then answer is marked correct, otherwise incorrect.

\section{Experimental Results}

\subsection{Question Generation}\label{question_generation_results}
We evaluate the generation quality of three models - \textit{T5-QG, BART-QG, BART-ML-QG} using standard language generation metrics: BLEU1-4 \cite{papineni2002bleu} and ROUGE-L \cite{rouge2004package} on the test set of 40 examples. The results are presented in \Cref{tab:qg_bleu_results}. We find BART outperforms T5 by 4 BLEU1 points, showing that BART is better suited for conditional generation. Also pre-training on MLQuestions dataset \cite{kulshreshtha2021back} increases BLEU1 by 1.5 absolute points.

\begin{table}[ht]
\begin{minipage}[b]{0.6\linewidth}
\centering
\begin{tabular}{c|ccccc}
    \toprule
        \textbf{Model} & B1 & B2 & B3 & B4 & R \\
        \hline
        T5-QG & 30.4 & 18.0 & 11.9 & 7.5 & 30.7 \\
        BART-QG & 34.5 & 24.5 & 17.1 & 12.1 & 39.3 \\
        \textbf{BART-ML-QG} & \textbf{36.1} & \textbf{24.7} & \textbf{19.6} & \textbf{12.2} & \textbf{39.7} \\
    \bottomrule
    \end{tabular}
    \caption{Results of Question Generation Models on \\\hspace{\textwidth}standard language evaluation metrics.}
    \label{tab:qg_bleu_results}
\end{minipage}\hfill
\begin{minipage}[b]{0.4\linewidth}
\centering
\includegraphics[scale=0.175]{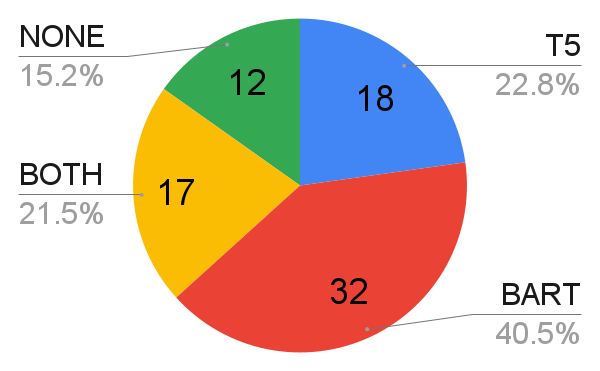}
\captionof{figure}{Comparing question quality of T5  \\\hspace{\textwidth}with BART based on annotated 80 questions.}
\label{fig:bartvst5}
\end{minipage}
\end{table}

\subsection{Question Re-ranking}

For question re-ranking, we experiment using different combinations of features described in \Cref{question_reranking} to predict usefulness score of generated question -
\begin{enumerate}
    \item \textbf{Mean Baseline:} This baseline simply outputs the usefulness as the average of all usefulness output in training set.
    \item \textbf{Linguistic:} Here we only use the four linguistic features - \textit{well-formedness, fluency, model confidence, question type score} as features for the question re-ranker.
    \item \textbf{SBERT:} Here we only use the 768-dimensional SBERT embedding features.
    \item \textbf{Ling-SBERT:} In this model we concatenate SBERT sentence embeddings with four linguistic features to train our question re-ranker.
\end{enumerate}
For each model we measure standard regression evaluation metrics - Mean Squared Error (MSE), Mean Absolute Error (MAE), Pearson Correlation (PCR). We also measure usefulness metric for each model. To compute usefulness, the re-ranker model predicts usefulness for each of the $k=3$ questions of the same given reference solution from the test set. Then the gold (actual) usefulness label for question achieving highest score is averaged across all reference solutions in test set. The results are presented in \Cref{tab:question_reranking_results}. We find that Ling-SBERT outperforms all other models for all metrics. More importantly, it improves the usefulness rating from 3.42 to 4.01. This means incorporating question re-ranking improves the actual usefulness of question generation by 0.5 on average!

\begin{table}
    \footnotesize
    \begin{tabular}{c|cccc}
    \toprule
        \textbf{Model} & MSE & MAE & PCR & Usefulness \\
        \hline
        Mean Baseline & 2.20 & 1.32 & - & 3.42 \\
        SBERT & 1.74 & 1.16 & 0.38 & 3.96 \\
        Linguistic & 1.78 & 1.16 & 0.33 & 3.85 \\
        \textbf{Ling-SBERT} & \textbf{1.72} & \textbf{1.15} & \textbf{0.40} & \textbf{4.01} \\
    \bottomrule
    \end{tabular}
    \caption{Results of Question Re-ranking.}
    \label{tab:question_reranking_results}
\end{table}

\subsection{Human Evaluation}
We manually compare the question quality generated by \textit{T5-QG} and \textit{BART-ML-QG} by generating questions for 80 randomly sampled reference solutions. The annotators compare both questions and provide one of the four labels - T5 (meaning T5 question is more useful than BART), BART (BART question is more useful), BOTH (both are equally good), and NONE (neither is a good question). The results from \Cref{fig:bartvst5} indicate that BART model is the clear winner, which is also supported by the superior BLEU scores.

Based on results on question generation, re-ranking and human evaluation we use the BART-ML-QG model for generation, and the Ling-SBERT model for re-ranking.

\subsection{Student Learning Gains}
After integrating our models in Korbit ITS, we collect around 146 distinct student interactions with feedback system for 550 exercises and measure student learning gains. The student learning gain is defined as the percentage of times a student answer is labelled correctly by the solution checker after they have received a given feedback. 
We compare our \textit{Personalized Question-based Feedback} with both simple and strong baselines:
\begin{itemize}
    \item \textbf{Minimal Feedback Baseline:} Here the system will simply tell the student that their solution is incorrect and they should try again.
    \item \textbf{Personalized Human Feedback Baseline:} For every exercise, Korbit already has several hints manually crafted by course designers. To select the best hint from the ones available, the ITS uses a personalized ML model by looking at student performance and responses on the exercise \cite{kochmar2021automated}. This personalization is used only during hint selection, and \textit{not} during hint generation itself (which is manual).
    \item \textbf{Personalized Non-question Feedback:} In this model after informing error type using cause-effect classifier, we reveal a part of the answer rather than asking a question. For e.g. if the student answers \textit{`Its a regression task because outputs are continuous'}, we show the hint as \textit{`"Its a regression task" is incorrect. Observe that outputs are discrete'} and ask the student to try again.
\end{itemize}

We present results of student learning gains in \Cref{tab:student_learning_gains}. The `First Attempt' column indicates entries in which the student tried only once previously, while `All Attempts' considers learning gains across student's all attempts. Our experiments show that our \textit{Personalized Question-based Feedback} model outperforms all models.

The \textit{Non-question Feedback} model improves over \textit{minimal feedback baseline} by 18\%, because it additionally informs about correct and incorrect/missing components. However, it cannot tell the student how to correct the incorrect/missing part. Our \textit{Question-based Feedback} model further improves over it by 26\%. This shows that asking questions about missing/incorrect parts is the key to help students improve their answers.

For all models, we observe that learning gains for `First Attempt' are more than `All Attempts'. This is likely because students who require many hints to solve an exercise may have knowledge gaps to solve exercises.

We find that most frequent student error is \textit{`incorrect cause incorrect effect'} followed by \textit{`missing cause correct effect'}. The error type \textit{`Correct cause incorrect effect'} occurs rarely as students usually know the main answer if they know the explanation behind it.

\begin{table}[]
    \centering
    \begin{tabular}{c|c|c}
    \toprule
    \multirow{2}{*}{} & \multicolumn{2}{c}{\textbf{Average Learning Gain (\%)}} \\
    \textbf{Model} & First Attempt & All Attempts \\
    \hline
    Minimal Feedback Baseline & 22.58 ± 14.72 & 21.74 ± 11.92 \\
    Personalized Human Feedback Baseline & 31.25 ± 16.06 & 30.43 ± 13.3 \\
    Personalized Non-question Feedback & 41.67 ± 19.72 & 34.38 ± 16.46 \\
    Personalized Question-based Feedback & \textbf{66.67 ± 16.87} & \textbf{52.27 ± 14.76} \\
    \bottomrule
    \end{tabular}
    \caption{Student learning gains on the Korbit ITS at 95\% confidence intervals.}
    \label{tab:student_learning_gains}
\end{table}

\section{Improving Generative Question Answering using Feedback Intervention}
Will a student having access to a feedback generation to correct it's mistakes during training perform better than another student without the feedback system support? Assume Student $S_A$ and $S_B$ are being taught by instructors $I_A$ and $I_B$. $I_A$ trains $S_A$ by  showing the answer for many questions. While $I_B$ trains $S_B$ by showing answers for questions, as well as sending \textit{personalized corrective feedback} when student answers question incorrectly. During test time, both students get same question paper without access to any feedback.\\ We simulate this behaviour by replacing student by QA model and teacher by hint model:
\begin{enumerate}
    \item Train baseline QA model $\theta_{QA}$ to generate reference solution from question.
    \item Generate machine (student) answers for questions in training data using $\theta_{QA}$. Generate  hints using our feedback system described previously for machine generated (incorrect) answer.
    \item Train hint generator $\theta_{HG}$ to generate these hints from question \& machine answer.
    \item Train hint-assisted QA model $\theta_{HQA}$ to generate answer from question and hint text generated by $\theta_{HG}$.
    \item During inference, first generate machine answer using $\theta_{QA}$. Next generate hint using $\theta_{HG}$ then generate final answer using $\theta_{HQA}$.
\end{enumerate}

The full algorithm is described in \ref{algo:hint_generation_improves_QA}. In principle we are first generating intermediate hint (part of answer) and then using it to generate the full answer. Similar inductive bias to learn the output in parts has been show to improve models in QA \cite{lamm2021qed} and QG \cite{kang2019let}.

\begin{algorithm}[t]
\begin{algorithmic}[1]
\footnotesize
\Require {
QA Data $\mathcal{D}_{QA} \equiv \{(q^i, a^i)\}_{i=1}^{m}$, Personalized Hint Generator $\mathcal{H}$
}
\Ensure Hint assisted QA model $\boldsymbol{\theta_{HQA}}$
\State $\boldsymbol{\theta_{QA}} \leftarrow $ Train on $\mathcal{D}_{QA}$ \Comment{Vanilla QA model}
\State $\mathcal{D}_{HG} \leftarrow [\ ]$ \Comment{Synthetic data for $\boldsymbol{\theta_{HG}}$}
\For{$q, a \in \mathcal{D}_{QA}$}
\State Generate machine answer $\hat{a} = \boldsymbol{\theta_{QA}}(q)$
\State Generate personalized hint $h = \mathcal{H}(q,\hat{a},a)$
\State add $(q,\hat{a},h)$ to $\mathcal{D}_{HG}$
\EndFor
\State $\boldsymbol{\theta_{HG}} \leftarrow $ Train on $\mathcal{D}_{HG}$ to generate $h$ from $(q,\hat{a})$
\State $\mathcal{D}_{HQA} \leftarrow [\ ]$ \Comment{Synthetic data for $\boldsymbol{\theta_{HQA}}$}
\For{$q, a \in \mathcal{D}_{QA}$}
\State Generate machine answer $\hat{a} = \boldsymbol{\theta_{QA}}(q)$
\State Generate hint $\hat{h} = \boldsymbol{\theta_{HG}}(q)$
\State add $(q,\hat{h},a)$ to $\mathcal{D}_{HQA}$
\EndFor
\State $\boldsymbol{\theta_{HQA}} \leftarrow $ Train on $\mathcal{D}_{HQA}$ to generate $a$ from $(q,\hat{h})$
\end{algorithmic}
\caption{Improving Generative QA using Personalized Feedback Generation}
\label{algo:hint_generation_improves_QA}
\end{algorithm}

\subsection{Hint-Answer Entailment Consistency}
In the above model, it is logical to expect the generated hint and model answer should be consistent with each other i.e. \textit{machine answer should \textbf{entail} model hint}. How can we enforce this inductive bias in the model? During inference, we generate $k=3$ model answers and measure the entailment probability of each answer to model generated hint using entailment probability of RoBERTa model \footnote{\url{https://huggingface.co/ynie/roberta-large-snli_mnli_fever_anli_R1_R2_R3-nli}} trained on multiple entailment datasets \cite{nie2019adversarial}. We pick the model answer having the highest entailment probability.

\subsection{Experiments and Results}
We use BART to train $\theta_{QA}, \theta_{HG}, \text{and\ } \theta_{HQA}$. Refer to \ref{qg_model_training_details} for model training details. Since there exists no generative cause-effect QA dataset to the best of our knowledge, we use Korbit dataset of 550 exercises and reference solutions. We split the data into 400 train, 50 validation and 100 test examples and measure BLEU and ROUGE metrics.

Experimental results presented in \Cref{tab:hint_assisted_qa} demonstrate that Hint-assisted QA system is superior to Vanilla-QA model by 1 ROUGE point, and enforcing hint-answer entailment further boosts ROUGE by up to 1.5 points.
Although the improvements are marginal, note that the task itself is hard as the training data is limited.

\begin{table}[]
    \centering
    \begin{tabular}{c|ccccc}
    \toprule
    \textbf{Models} & BLEU1 & BLEU2 & BLEU3 & BLEU4 & ROUGE-L \\
    \hline
    \textit{Vanilla-QA} & 24.57 & 14.89 & 10.70 & 8.27 & 29.68 \\
    \textit{Hint-assisted QA} & 25.16 & 15.07 & 11.56 & 9.37 & 30.63 \\
    \textbf{\textit{Hint+Entailment}} & \textbf{25.54} & \textbf{16.05} & \textbf{12.19} & \textbf{9.57} & \textbf{31.35} \\
    \bottomrule
    \end{tabular}
    \caption{Results on improving Generative Question Answering Using Hint Intervention}
    \label{tab:hint_assisted_qa}
\end{table}

\section{Related Work}

\paragraph{Feedback Generation}

Previous research on dialogue-based ITS similar to Korbit investigated various aspects of automated feedback generation and adaptation~\citep{benzmuller2007natural,makatchev2011representation,stamper2013}. In particular,~\citet{stamper2013} investigated ways to augment their Deep Thought logic tutor with a Hint Factory that generated data-driven, context-specific hints for an ITS. The hints were effective in promoting learning, however, their approach mostly focused on the automated detection of the best hint sequence among hints consisting of logic rules, whereas our work focuses on methods of hint generation in natural language. The most similar work to ours is that \citet{grenander2021deep}, who also generate personalized feedback based on cause–effect analysis, but do not use questions in their generated feedback, hence their feedback does not reveal any hint about correct answer. 

\paragraph{Question Generation}Previous research has focused on training neural Seq2Seq models \cite{du2017learning,zhao2018paragraph,klein2019learning} on supervised full QA datasets such as SQuAD \cite{rajpurkar2016squad}. QG in a few-shot setting under limited data has also been explored recently for multi-hop QG \cite{yu2020low, huang2021latent}.

\citet{chen2018learningq} create a large-scale Educational QG dataset from KhanAcademy and TED-Ed data sources as a learning and assessment tools for students. \citet{kulshreshtha2021back} also release a QG dataset comprising of data-science questions to promote research in domain adaptation. Unlike our questions, the questions in \citet{chen2018learningq,kulshreshtha2021back} are static and not personalized to the student. A recent work by \citet{srivastava2021question} generates personalized questions according to the student's level by proposing a difficulty-controllable QG model. To the best of our knowledge, we are the first to use QG in an education context with real student interaction data.

\paragraph{Improving Question Answering using Hints}is not yet studied clearly in NLP paradigm. A related work by \citet{lamm2021qed} proposes the use of \textit{explanations} for an answer to improve Question Answering. They annotate a dataset of 8,991 QED explanations and use it to learn joint QA and explanation generation. Their explanations however are very different from our hints as they are non-personalized (fixed for a given question/answer). 

\section{Conclusion and Future Work}
We show how can we provide personalized feedback to students in an ITS by combining rule-based models such as cause-effect extraction with deep-learning models such as few-shot Question generation and semantic similarity. Our approach identifies correct and incorrect/missing components in student answers using cause-effect analysis and BERT Transformer. The few-shot Question Generation and re-ranker model then generates questions to help improve student answer. Our model vastly outperforms both simple and strong baselines on student learning gains by a large margin on the Korbit ITS. 

One area of future research is to design personalizing feedback for non cause-effect exercises. Another idea is to show multiple feedback to students and have them evaluate it either explicitly or implicitly by trying to answer the question-based feedback. This training signal can be used to further improve the feedback model using active learning.

\appendix
\section{Appendix - Question Generation Model Training Details}\label{qg_model_training_details}
All three models - \textit{T5-QG, BART-QG, BART-ML-QG} are trained for 5 epochs with learning rate of $1e-5$ and batch size of 8. For optimization we use Adam \cite{kingma2014adam} with $\beta_1 = 0.9, \beta_2 = 0.999$. The input and output sequence length is padded to 512 and 150 tokens respectively. For generation we use beam search decoding \cite{freitag2017beam} with number of beams set to 3. The initial checkpoint for the models can be found at -  T5-QG\footnote{\url{https://huggingface.co/docs/transformers/v4.18.0/en/model_doc/t5\#transformers.T5ForConditionalGeneration}}, BART-QG\footnote{\url{https://huggingface.co/docs/transformers/v4.18.0/en/model_doc/bart\#transformers.BartForConditionalGeneration} \label{BART-checkpoint}}, BART-ML-QG\footnote{\url{https://huggingface.co/McGill-NLP/bart-qg-mlquestions-backtraining}}. The hint-assisted QA models - $\theta_{QA}, \theta_{HG}, \theta_{HQA}$ are trained using same configurations and vanilla BART checkpoint.

\bibliography{ios-book-article}

\begin{thebibliography}{41}
\providecommand{\natexlab}[1]{#1}
\providecommand{\url}[1]{\texttt{#1}}
\expandafter\ifx\csname urlstyle\endcsname\relax
  \providecommand{\doi}[1]{doi: #1}\else
  \providecommand{\doi}{doi: \begingroup \urlstyle{rm}\Url}\fi

\bibitem[Benzm{\"u}ller et~al.(2007)Benzm{\"u}ller, Horacek, Kruijff-Korbayova,
  Pinkal, Siekmann, and Wolska]{benzmuller2007natural}
Christoph Benzm{\"u}ller, Helmut Horacek, Ivana Kruijff-Korbayova, Manfred
  Pinkal, J{\"o}rg Siekmann, and Magdalena Wolska.
\newblock Natural language dialog with a tutor system for mathematical proofs.
\newblock In \emph{Cognitive Systems}, pages 1--14. Springer, 2007.

\bibitem[Blayney and Freeman(2004)]{blayney2004automated}
Paul Blayney and Mark Freeman.
\newblock Automated formative feedback and summative assessment using
  individualised spreadsheet assignments.
\newblock \emph{Australasian Journal of Educational Technology}, 20\penalty0
  (2), 2004.

\bibitem[Brown et~al.()Brown, Mann, Ryder, Subbiah, Kaplan, Dhariwal,
  Neelakantan, Shyam, Sastry, Askell, et~al.]{brown2020language}
Tom Brown, Benjamin Mann, Nick Ryder, Melanie Subbiah, Jared~D Kaplan, Prafulla
  Dhariwal, Arvind Neelakantan, Pranav Shyam, Girish Sastry, Amanda Askell,
  et~al.
\newblock Language models are few-shot learners.
\newblock \emph{NIPS 2020}.

\bibitem[Cao et~al.(2016)Cao, Sun, and Zhuge]{cao2016role}
Mengyun Cao, Xiaoping Sun, and Hai Zhuge.
\newblock The role of cause-effect link within scientific paper.
\newblock In \emph{2016 12th International Conference on Semantics, Knowledge
  and Grids (SKG)}, pages 32--39. IEEE, 2016.

\bibitem[Chen et~al.(2018)Chen, Yang, Hauff, and Houben]{chen2018learningq}
Guanliang Chen, Jie Yang, Claudia Hauff, and Geert-Jan Houben.
\newblock Learningq: a large-scale dataset for educational question generation.
\newblock In \emph{Twelfth International AAAI Conference on Web and Social
  Media}, 2018.

\bibitem[Devlin et~al.()Devlin, Chang, Lee, and Toutanova]{devlin2019bert}
Jacob Devlin, Ming-Wei Chang, Kenton Lee, and Kristina Toutanova.
\newblock Bert: Pre-training of deep bidirectional transformers for language
  understanding.
\newblock In \emph{NAACL 2019}, pages 4171--4186.

\bibitem[Dong et~al.(2019)Dong, Yang, Wang, Wei, Liu, Wang, Gao, Zhou, and
  Hon]{dong2019unified}
Li~Dong, Nan Yang, Wenhui Wang, Furu Wei, Xiaodong Liu, Yu~Wang, Jianfeng Gao,
  Ming Zhou, and Hsiao-Wuen Hon.
\newblock Unified language model pre-training for natural language
  understanding and generation.
\newblock \emph{NIPS}, 2019.

\bibitem[Du et~al.(2017)Du, Shao, and Cardie]{du2017learning}
Xinya Du, Junru Shao, and Claire Cardie.
\newblock Learning to ask: Neural question generation for reading
  comprehension.
\newblock In \emph{ACL}, 2017.

\bibitem[Faruqui and Das(2018)]{faruqui2018identifying}
Manaal Faruqui and Dipanjan Das.
\newblock Identifying well-formed natural language questions.
\newblock In \emph{EMNLP}, 2018.

\bibitem[Freitag and Al-Onaizan(2017)]{freitag2017beam}
Markus Freitag and Yaser Al-Onaizan.
\newblock Beam search strategies for neural machine translation.
\newblock In \emph{Proceedings of the First Workshop on Neural Machine
  Translation}, pages 56--60, 2017.

\bibitem[Grenander et~al.(2021)Grenander, Belfer, Kochmar, Serban, St-Hilaire,
  and Cheung]{grenander2021deep}
Matt Grenander, Robert Belfer, Ekaterina Kochmar, Iulian~V Serban,
  Fran{\c{c}}ois St-Hilaire, and Jackie~CK Cheung.
\newblock Deep discourse analysis for generating personalized feedback in
  intelligent tutor systems.
\newblock In \emph{The 11th Symposium on Educational Advances in Artificial
  Intelligence}, 2021.

\bibitem[Huang et~al.(2021)Huang, Qi, Sun, and Zhang]{huang2021latent}
Xinting Huang, Jianzhong Qi, Yu~Sun, and Rui Zhang.
\newblock Latent reasoning for low-resource question generation.
\newblock In \emph{Findings of ACL}, 2021.

\bibitem[Kang et~al.(2019)Kang, San~Roman, and Myaeng]{kang2019let}
Junmo Kang, Haritz~Puerto San~Roman, and Sung-Hyon Myaeng.
\newblock Let me know what to ask: Interrogative-word-aware question
  generation.
\newblock In \emph{Proceedings of the 2nd Workshop on Machine Reading for
  Question Answering}, pages 163--171, 2019.

\bibitem[Kingma and Ba(2014)]{kingma2014adam}
Diederik~P Kingma and Jimmy Ba.
\newblock Adam: A method for stochastic optimization.
\newblock \emph{arXiv preprint arXiv:1412.6980}, 2014.

\bibitem[Klein and Nabi(2019)]{klein2019learning}
Tassilo Klein and Moin Nabi.
\newblock Learning to answer by learning to ask: Getting the best of gpt-2 and
  bert worlds.
\newblock \emph{arXiv preprint arXiv:1911.02365}, 2019.

\bibitem[Kochmar et~al.(2021)Kochmar, Vu, Belfer, Gupta, Serban, and
  Pineau]{kochmar2021automated}
Ekaterina Kochmar, Dung~Do Vu, Robert Belfer, Varun Gupta, Iulian~Vlad Serban,
  and Joelle Pineau.
\newblock Automated data-driven generation of personalized pedagogical
  interventions in intelligent tutoring systems.
\newblock \emph{International Journal of Artificial Intelligence in Education},
  2021.

\bibitem[Kulshreshtha et~al.(2021)Kulshreshtha, Belfer, Vlad~Serban, and
  Reddy]{kulshreshtha2021back}
Devang Kulshreshtha, Robert Belfer, Iulian Vlad~Serban, and Siva Reddy.
\newblock Back-training excels self-training at unsupervised domain adaptation
  of question generation and passage retrieval.
\newblock \emph{arXiv e-prints}, pages arXiv--2104, 2021.

\bibitem[Lamm et~al.(2021)Lamm, Palomaki, Alberti, Andor, Choi, Soares, and
  Collins]{lamm2021qed}
Matthew Lamm, Jennimaria Palomaki, Chris Alberti, Daniel Andor, Eunsol Choi,
  Livio~Baldini Soares, and Michael Collins.
\newblock Qed: A framework and dataset for explanations in question answering.
\newblock \emph{Transactions of the Association for Computational Linguistics},
  9:\penalty0 790--806, 2021.

\bibitem[Landis and Koch(1977)]{landis1977measurement}
J~Richard Landis and Gary~G Koch.
\newblock The measurement of observer agreement for categorical data.
\newblock \emph{biometrics}, pages 159--174, 1977.

\bibitem[Lewis et~al.(2020)Lewis, Liu, Goyal, Ghazvininejad, Mohamed, Levy,
  Stoyanov, and Zettlemoyer]{lewis2020bart}
Mike Lewis, Yinhan Liu, Naman Goyal, Marjan Ghazvininejad, Abdelrahman Mohamed,
  Omer Levy, Veselin Stoyanov, and Luke Zettlemoyer.
\newblock Bart: Denoising sequence-to-sequence pre-training for natural
  language generation, translation, and comprehension.
\newblock In \emph{ACL}, 2020.

\bibitem[Li et~al.(2019)Li, Lei, Qin, and Wang]{li2019deep}
Siyao Li, Deren Lei, Pengda Qin, and William~Yang Wang.
\newblock Deep reinforcement learning with distributional semantic rewards for
  abstractive summarization.
\newblock \emph{arXiv preprint arXiv:1909.00141}, 2019.

\bibitem[Liu et~al.(2016)Liu, Li, Xu, and Liu]{liu2016automated}
Ming Liu, Yi~Li, Weiwei Xu, and Li~Liu.
\newblock Automated essay feedback generation and its impact on revision.
\newblock \emph{IEEE Transactions on Learning Technologies}, 10\penalty0
  (4):\penalty0 502--513, 2016.

\bibitem[Makatchev et~al.(2011)Makatchev, Jordan, Pappuswamy, and
  VanLehn]{makatchev2011representation}
Maxim Makatchev, Pamela~W Jordan, Umarani Pappuswamy, and Kurt VanLehn.
\newblock Representation and reasoning for deeper natural language
  understanding in a physics tutoring system.
\newblock \emph{AAAI}, 2011.

\bibitem[McBroom et~al.(2021)McBroom, Koprinska, and Yacef]{mcbroom2021survey}
Jessica McBroom, Irena Koprinska, and Kalina Yacef.
\newblock A survey of automated programming hint generation: The hints
  framework.
\newblock \emph{ACM Computing Surveys (CSUR)}, 54\penalty0 (8):\penalty0 1--27,
  2021.

\bibitem[Nie et~al.(2019)Nie, Williams, Dinan, Bansal, Weston, and
  Kiela]{nie2019adversarial}
Yixin Nie, Adina Williams, Emily Dinan, Mohit Bansal, Jason Weston, and Douwe
  Kiela.
\newblock Adversarial nli: A new benchmark for natural language understanding.
\newblock \emph{arXiv preprint arXiv:1910.14599}, 2019.

\bibitem[Oberm{\"u}ller et~al.(2021)Oberm{\"u}ller, Heuer, and
  Fraser]{obermuller2021guiding}
Florian Oberm{\"u}ller, Ute Heuer, and Gordon Fraser.
\newblock Guiding next-step hint generation using automated tests.
\newblock In \emph{Proceedings of the 26th ACM Conference on Innovation and
  Technology in Computer Science Education V. 1}, pages 220--226, 2021.

\bibitem[Papineni et~al.(2002)Papineni, Roukos, Ward, and
  Zhu]{papineni2002bleu}
Kishore Papineni, Salim Roukos, Todd Ward, and Wei-Jing Zhu.
\newblock Bleu: a method for automatic evaluation of machine translation.
\newblock In \emph{ACL}, 2002.

\bibitem[Raffel et~al.(2020)Raffel, Shazeer, Roberts, Lee, Narang, Matena,
  Zhou, Li, and Liu]{raffel2020exploring}
Colin Raffel, Noam Shazeer, Adam Roberts, Katherine Lee, Sharan Narang, Michael
  Matena, Yanqi Zhou, Wei Li, and Peter~J Liu.
\newblock Exploring the limits of transfer learning with a unified text-to-text
  transformer.
\newblock \emph{Journal of Machine Learning Research}, 21:\penalty0 1--67,
  2020.

\bibitem[Rajpurkar et~al.(2016)Rajpurkar, Zhang, Lopyrev, and
  Liang]{rajpurkar2016squad}
Pranav Rajpurkar, Jian Zhang, Konstantin Lopyrev, and Percy Liang.
\newblock Squad: 100,000+ questions for machine comprehension of text.
\newblock \emph{arXiv preprint arXiv:1606.05250}, 2016.

\bibitem[Reimers and Gurevych(2019)]{reimers2019sentence}
Nils Reimers and Iryna Gurevych.
\newblock Sentence-bert: Sentence embeddings using siamese bert-networks.
\newblock In \emph{EMNLP}, 2019.

\bibitem[Rothe et~al.(2021)Rothe, Maynez, and Narayan]{rothe2021thorough}
Sascha Rothe, Joshua Maynez, and Shashi Narayan.
\newblock A thorough evaluation of task-specific pretraining for summarization.
\newblock In \emph{EMNLP}, 2021.

\bibitem[ROUGE(2004)]{rouge2004package}
Lin~CY ROUGE.
\newblock A package for automatic evaluation of summaries.
\newblock In \emph{Proceedings of Workshop on Text Summarization of ACL,
  Spain}, 2004.

\bibitem[Srivastava and Goodman(2021)]{srivastava2021question}
Megha Srivastava and Noah Goodman.
\newblock Question generation for adaptive education.
\newblock In \emph{ACL}, 2021.

\bibitem[St-Hilaire et~al.(2021)St-Hilaire, Burns, Belfer, Shayan, Smofsky, Vu,
  Frau, Potochny, Faraji, Pavero, et~al.]{st2021comparative}
Francois St-Hilaire, Nathan Burns, Robert Belfer, Muhammad Shayan, Ariella
  Smofsky, Dung~Do Vu, Antoine Frau, Joseph Potochny, Farid Faraji, Vincent
  Pavero, et~al.
\newblock A comparative study of learning outcomes for online learning
  platforms.
\newblock In \emph{International Conference on Artificial Intelligence in
  Education}, pages 331--337. Springer, 2021.

\bibitem[St-Hilaire et~al.(2022)St-Hilaire, Vu, Frau, Burns, Faraji, Potochny,
  Robert, Roussel, Zheng, Glazier, et~al.]{st2022new}
Francois St-Hilaire, Dung~Do Vu, Antoine Frau, Nathan Burns, Farid Faraji,
  Joseph Potochny, Stephane Robert, Arnaud Roussel, Selene Zheng, Taylor
  Glazier, et~al.
\newblock A new era: Intelligent tutoring systems will transform online
  learning for millions.
\newblock \emph{arXiv preprint arXiv:2203.03724}, 2022.

\bibitem[Stamper et~al.(2013)Stamper, Eagle, Barnes, and Croy]{stamper2013}
John~C. Stamper, Michael Eagle, Tiffany Barnes, and Marvin Croy.
\newblock {Experimental Evaluation of Automatic Hint Generation for Logic
  Tutor}.
\newblock \emph{International Journal of Artificial Intelligence in Education},
  22\penalty0 (1-2):\penalty0 3--17, 2013.

\bibitem[Unanue et~al.(2021)Unanue, Parnell, and Piccardi]{unanue2021berttune}
Inigo~Jauregi Unanue, Jacob Parnell, and Massimo Piccardi.
\newblock Berttune: Fine-tuning neural machine translation with bertscore.
\newblock \emph{arXiv preprint arXiv:2106.02208}, 2021.

\bibitem[Wenger(1987)]{Wenger}
\'{E}tienne Wenger.
\newblock \emph{{Artificial Intelligence and Tutoring Systems}}.
\newblock Los Altos, CA: Morgan Kaufmann, 1987.

\bibitem[Yu et~al.(2020)Yu, Liu, Qiu, Su, Wang, Quan, and Yin]{yu2020low}
Jianxing Yu, Wei Liu, Shuang Qiu, Qinliang Su, Kai Wang, Xiaojun Quan, and Jian
  Yin.
\newblock Low-resource generation of multi-hop reasoning questions.
\newblock In \emph{ACL}, 2020.

\bibitem[Zhang et~al.(2019)Zhang, Kishore, Wu, Weinberger, and
  Artzi]{zhang2019bertscore}
Tianyi Zhang, Varsha Kishore, Felix Wu, Kilian~Q Weinberger, and Yoav Artzi.
\newblock Bertscore: Evaluating text generation with bert.
\newblock \emph{arXiv preprint arXiv:1904.09675}, 2019.

\bibitem[Zhao et~al.(2018)Zhao, Ni, Ding, and Ke]{zhao2018paragraph}
Yao Zhao, Xiaochuan Ni, Yuanyuan Ding, and Qifa Ke.
\newblock Paragraph-level neural question generation with maxout pointer and
  gated self-attention networks.
\newblock In \emph{EMNLP}, 2018.

\end{thebibliography}
\bibliographystyle{plainnat}

\end{document}